# Why Molière most likely did write his plays

Florian Cafiero[1]* and Jean-Baptiste Camps[2]*

As for Shakespeare, a hard-fought debate has emerged about Molière, a supposedly uneducated actor who, according to some, could not have written the masterpieces attributed to him. In the past decades, the century-old thesis according to which Pierre Corneille would be their actual author has become popular, mostly because of new works in computational linguistics. These results are reassessed here through state-of-the-art attribution methods. We study a corpus of comedies in verse by major authors of Molière and Corneille's time. Analysis of lexicon, rhymes, word forms, affixes, morphosyntactic sequences, and function words do not give any clue that another author among the major playwrights of the time would have written the plays signed under the name Molière.

## INTRODUCTION

Even if some have argued it dates back to the 17th century (1), the Molière authorship question mostly gained public attention during the first half of the 20th century. In 1919, the French writer Pierre Louÿs claimed that Pierre Corneille was the ghostwriter of most famous plays attributed to Molière (2, 3). The late blooming of Molière's talent, his purported lack of education and culture, his busy agenda, and the lack of manuscripts are among the arguments that triggered a century-long debate (1). Systematic objections to these assertions have been provided (4). Yet, the sparsity of available archives has so far prevented the debate from ending.

Quantitative analysis of the texts revived the controversy (5, 6). C. and D. Labbé defined an "intertextual distance," measuring the difference in lexicon between texts. After measuring the distances observed between various literary texts, they determine a threshold, under which two plays should be considered as written by the same author. According to their criterion, texts by Molière and P. Corneille would have been written by the same person. The Molière case would not be an exception, but the sign of a wide system of ghostwriting. According to D. Labbé (7), most plays of the mid-17th century would have been signed under the name of "poet actors" (comédiens poètes), prominent performers publicly presented as the author of the plays in which they star. More generally, authorship would not have been a well-respected concept. According to D. Labbé's own estimation (8), around 90% of the comedies and half of the tragedies of that time would have been signed by a "poet actor," acting as figurehead, and not by their actual author.

These conclusions were given a large media exposure (9, 10) and fostered a vigorous debate in academia. Bernet (11) found the intertextual distance separating Molière and Corneille to be standard for plays of that time. C. and D. Labbé (12) objected that Bernet might not have respected their method and that the small distance observed between most of these authors' works could mean that P. Corneille and his brother Thomas could have written most plays in Bernet's corpus as well—the authors chosen by Bernet being, in their opinion, only figureheads and not proper playwrights. Schöch (13) asserted that C. and D. Labbé's method would artificially smooth differences between authors. Using Euclidean distance increases the weight of frequent lemmas, and working on lemmatized texts tends to lower the disparity between the observed frequencies of various forms.

Other works have then added to this literature. Vonfelt (14) proposed a variant of C. and D. Labbé's intertextual distance, focusing on 1-grams of characters. The lack of selection or treatment on the texts is hard to defend. Marusenko and Rodionova (15) tried to attribute 13 comedies in verse signed by Molière. These plays are compared to two sets of comedies in verse, one by P. Quinault and the other by P. Corneille, according to five grammatical criteria. A weighted Euclidean distance between the alleged plays by Molière and each of the comparison groups is then computed. If a play is closer to one of the groups, it is attributed to its author. In case of tie, it is not attributed. A "deterministic" algorithm attributes six plays to P. Corneille, while a "probabilistic" one attributes four remaining plays to Corneille and one to Quinault. Only two plays remain unattributed. This methodology raises many questions. First, the authors do not benchmark their procedure, making it impossible to evaluate their method's empirical performance. Then, the corpora of plays by Quinault and Corneille are extremely uneven: 3 plays by Quinault, written over a decade, versus 11 plays by P. Corneille, written over more than 40 years. This could lead to numerous biases in the procedure. Last, the logic of the procedure itself is disturbing: Observing that a text is closer to Corneille than to Quinault does not mean it is written by Corneille. More qualitative approaches have also been proposed to apprehend this problem, convincingly spotting minor differences in versification (16) or in lexicon (17), without being able to make any decisive argument.

The Molière controversy is actually intricate to solve through textual analysis, as it raises many specific concerns compared to most authorship attribution problems. The generalized doubt about official authors of comedies, raised by D. Labbé, prevents us from taking any information regarding authors for granted. It is thus logically impossible, for instance, to use supervised methods to identify the styles of at least some of the authors of the time. Another problem comes from the nature of the texts that we have to analyze. French theater at Molière's time, known as théâtre classique, is admittedly very homogeneous in style and topics (18). It is rigorously codified ("classical unities" of action, time and place, decency rules, etc.) and follows very strict versification rules. Most plays are adapted from Spanish (Calderòn, Lope de Vega, Moreto, Rojas), Italian (Barbieri, Piccolomini, Dalla Porta), or Ancient (Plaute) models. A same source could be an inspiration to different playwrights: Rotrou's Les Sosies and Molière's Amphitryon both derive from Plaute's Amphitryon; T. Corneille's Le Geôlier de soi-même and Scarron's Le Gardien de soi-même from Calderòn's El Alcaide de sí mismo, etc. This is without mentioning the numerous inspirations and borrowings between contemporary authors themselves. For instance, T. Corneille was notably inspired by his brother's Le Menteur and la Suite du Menteur when writing for instance *L'Amour à la mode* (*19*). Molière's plays borrow some expressions and situations from d'Ouville (*20*), Scarron, or the Corneille brothers (*4*). All this contributes to make the potential stylistic variations between authors remarkably tenuous, especially in comparison to even minor generic variations. We are thus compelled to be very strict when defining our final corpus. Last, the lack of manuscripts, extremely rarely available for plays of the 17th century, and nonexistent for P. Corneille or Molière's plays, forces us to consider punctuation and variations in spelling as irrelevant, as they could be the doing of the editors, and not of the authors themselves.

[1]LIED, Université Paris-Diderot/Université de Paris, CNRS, Paris, France. [2]Centre Jean-Mabillon, École nationale des chartes, Université Paris Sciences et Lettres, 65, rue de Richelieu, Paris 75002, France.
*Corresponding author. Email: florian.cafiero@polytechnique.edu (F.C.); jean-baptiste.camps@chartes.psl.eu (J.-B.C.)

One of the only characteristics easing this Molière problem is the number of historically plausible authors. In the Shakespeare dispute, a somewhat similar controversy, many candidates have been suggested to be the actual author: Edward de Vere, Earl of Oxford; Francis Bacon; Christopher Marlowe; William Stanley, Earl of Derby, attracted the most attention among the 80 authors at least once suggested as Shakespeare's ghostwriter. In Molière's case, only P. Corneille has ever been considered as a potential ghostwriter—his brother Thomas being, in rare cases, marginally involved. In that respect, our question is thus simpler to address: Is there any reason to suspect that P. Corneille or his brother would have written at least some of Molière's plays? Recent advances in authorship attribution methods for literary texts (*21*, *22*) and wider availability of properly edited and digitized texts now allow a more reliable answer to that question.

Here, we work on a corpus of plays from the same genre ("comédie"), all of them in verse, written by Molière, P. Corneille, T. Corneille, and other major authors of their time, prolific enough in the same genre to be compared to them. We systematically analyze their global lexicon, rhyme lexicon, word forms, affixes, morphosyntax, and function words.

We test here the two major theories challenging Molière's authorship. Following Wouters *et al.* (*23*), the first hypothesis (H1) states that Molière would have provided P. Corneille with drafts that the latter would then versify—with the help of his brother or not. Molière would have created the plots—which justifies him signing—but the versification, considered as a technical operation, and thus undeserving of explicit credit, would have been realized by P. Corneille (or his brother). If this hypothesis were true, then similarities in vocabulary could be noticeable in plays signed by Molière. But rhymes, function words, affixes, and morphosyntactic sequences should be the same as in other P. (or T.) Corneille's plays. The second hypothesis (H2), following D. Labbé's poet actor theories, asserts that Molière would have written nor the plots nor the verses of his plays. Molière would have only been a famous name, used like others at the time to help promote the play, to satisfy the main actor's/director's ego, and to conceal the name of the actual author, supposedly unwilling to be known as the writer of comedies. Topics chosen in Molière's plays, like for instance the *Précieuses Ridicules*, would have been closer to P. (or T.) Corneille's usual interests and would not reflect any influence by Molière. If this hypothesis were true, then all our indicators should show that there is no such thing as Molière's vocabulary or Molière's style, and Molière's plays should be confused with P. Corneille's plays according to each of our six criteria.

Our analysis disproves both theories and concludes that neither P. Corneille nor T. Corneille (and incidentally, nor any of the major authors tested here) could have written the plays signed under the name Molière. Without definitely proving that Molière's works are his own—which only historical evidence could do—disproving these alternative theories strongly substantiates the idea that Molière indeed wrote the masterpieces signed under his name.

## RESULTS
To measure similarities between texts, different sets of features can be quantified, be they lexical (lemmas, rhyme lemmas, word forms) or grammatical (affixes, part-of-speech, function words). For each of these six types of feature, texts are sorted according to a hierarchical clustering algorithm, following a procedure tested with success on a control corpus of comedies in verse, written after Molière and Corneille's death (see Materials and Methods). We first perform an exploratory analysis on a large corpus of plays of Molière's time. We then analyze a corpus of plays belonging to the same subgenre as major comedies signed by Molière.

## Features studied
### Lexicon
On the matter at hand, seminal studies by C. and D. Labbé focused on the frequencies of lemmas or canonical forms of the words (i.e., "to love" is the lemma of "loving," "loved," "(he) loves," etc.). These features are heavily dependent on a text's topic or literary genre. Studying them is helpful anyway in our context: If our H1 hypothesis were true, then plays by P. Corneille and Molière should differ according to this criterion but show similarities according to all other angles.

### Rhyme lexicon
The words chosen at the rhyme are carefully selected by the writer and are sometimes seen as specific to an author (*24*). Of course, words at the rhyme still depend on the topic of the play, and similarities in the rhymes vocabulary can easily be the result of intentional imitation or unconscious inspiration. Moreover, the sample of the lemmas used in rhyme position is a substantially smaller subset of the sample containing all lemmas from a text.

### Word forms
Raw (i.e., unlemmatized) word forms are sometimes deemed revealing of authorial style: While they are still heavily dependent on the content of the texts, the absence of lemmatization allows the preservation of some morphological information (inflection of nouns and verbs), which can be helpful in an inflected language such as French.

### Affixes
Character *n*-grams—that is, sequences of *n* contiguous characters—are often considered a very effective feature for authorship attribution (*25*). Their effectiveness seems to come from their ability to capture grammatical morphemes, especially prefixes and suffixes, as well as authorial punctuation when available (*26*). In the absence of authorial punctuation, we restrict our analysis of character *n*-grams to four types of "affixes": the three first or last character of the words of at least four characters, as well as the interword space with the two characters preceding or following it (*26*).

### Morphosyntactic sequences
Differences in morphosyntax are considered an important clue to determine the author of a text. One possible approach is to analyze the sequences of word classes or part of speech (POS). POS *n*-grams, in particular POS 3-grams (i.e., sequences of tags, such as "NOUN ADJECTIVE VERB"), proved to be effective criteria to discover the author of a text—outperformed only by function words (*27*).

### Function words
According to the literature of the past 3 decades, the analysis of function words is the most reliable method for literary authorship attribution (*25*, *28*). The underlying intuition is that function words are used mostly according to unconscious patterns and vary less according to the topics or the genre of the texts. Psycholinguistics studies have shown that function words are perceived by the readers on a less conscious level and are read faster than content words; they might also be chosen less consciously by the writers, while nonetheless being able to convey significant information on the speaker or writer (*25*, *29*). Shared by all writers, function words are also interesting from a statistical perspective, as the number of distinct function words in a language is small (e.g., around 100 in French), but their number of occurrences in a text is very high (*25*, *29*).

### Exploratory analysis

We first perform an exploratory analysis of a large sample of comedies in verse, using Fièvre's digital editions (*30*). This sample includes plays of at least 5000 words, for authors having written at least three comedies. It includes plays by 12 authors. Results are displayed in Fig. 1.

The highest agglomerative coefficient is obtained for the analysis of function words. In this analysis, all plays signed by Molière are clustered together, and plays are in vast majority clustered according to their alleged author. More generally, clustering according to all of our feature sets (Fig. 1, A, C, D, E, and F) but rhyme vocabulary (Fig. 1B) leads to a structure mainly congruent with the alleged authorship of the plays. Major works attributed to Molière are consistently clustered together, especially when examining more topic-independent features such as POS 3-grams or functions words (Fig. 1, E and F).

Yet, for lemmas and word forms (Fig. 1, A and C), the main subdivision in three clusters is not authorial but partly generic and partly length based. A first cluster consists of farce-like plays (e.g., Molière's *Sganarelle ou le Cocu imaginaire*, its imitation by Donneau de Visé, *La Cocue imaginaire*, Chevalier's plays…) and shorter plays such as *Les Fâcheux* or *Mélicerte*. The central cluster comprises regular comedies and, in particular, most works attributed to Molière. The last cluster includes a specific subgroup of heroic comedies, *Don Sanche*, *Pulchérie*, *Tite*, as well as *Dom Garcie de Navarre* and the *Illustres ennemis*. The features most specific to this cluster seem to differentiate it clearly from the rest of the corpus (see Materials and Methods).

Cases of common inspiration, sometimes on the verge of plagiarism, also yield some anomalies. Two homonymous plays, the *Mère coquette ou les amants brouillés*, published by Quinault (1665), then by Donneau de Visé (1666)—who claims the paternity of the subject—are clustered together. A similar phenomenon links Molière's *Sganarelle ou le cocu imaginaire* with (again) Donneau de Visé's imitation, *La Cocue imaginaire*. We also note that *les Soupçons sur les Apparences* is not consistently clustered between analyses. The attribution to d'Ouville of this particular work is uncertain, as it dates back to 17th century catalogs and raised suspicions even at the time (*31*). It could have been authored by an author outside of this corpus or coauthored by d'Ouville and someone else. A similar phenomenon is, for instance, observed for the *Ragotin* signed by La Fontaine but written in collaboration with Champmeslé, a comedian and author, whose contribution is still unclear.

These analyses show that the cluster containing Molière's alleged most important plays is one of the strongest and most distinct clusters, especially when we look at more topic or genre independent features, such as POS 3-grams or functions words (Fig. 1, E and F). Moreover, it is not mixed with the works of P. or T. Corneille.

Nonetheless, various phenomena seem to interfere with our analysis: variations due to subgenres, doubtful authorship, collaboration, plagiarism, variations in size of the authors' sample, etc. This urges us to build a more controlled subcorpus, focusing only on a less diverse set of plays and authors but comprising only works directly comparable to Molière's supposed masterpieces in the genre.

### Final analysis

To obtain a more readable and less biased result, we then lead a subgroup analysis. To avoid biases due to subgenres, we exclude *comédies héroïques* ("heroic comedies") and short farcesque comedies. To eliminate the noise added by many phenomena (coauthoring, plagiarism, uncertain attribution, etc.) irrelevant to the hypotheses we are testing, we choose to work only on five major authors of the time. This final corpus contains 37 plays by T. and P. Corneille, Molière, Rotrou, and Scarron. Results are shown in Fig. 2.

According to our six criteria, all plays by Molière belong to the same cluster. The purity of this cluster is 1 for our six analyses. The clustering with the highest agglomerative coefficient is again the one based on function words. It attributes 100% of the plays to their alleged author. Other criteria also attribute an overwhelming majority of plays to their alleged author (over 95% cluster purity). The same phenomenon explains this minimal difference between alleged authors and predicted author. T. Corneille's early plays and two plays by his brother (*Le Menteur* and *La suite du Menteur*) are, according to some criteria, mixed in the same cluster. Critics already noted that T. Corneille drew his inspiration (*19*) from these two plays. This strong similarity does not come as much of a surprise. It could reflect the strong influence the already famous P. Corneille had on his much younger brother at early stages of his career, either punctual collaboration, corrections, or advice given by P. Corneille. All other plays are always clustered according to their alleged author.

### DISCUSSION

Our study aimed to test two hypothesis. The first hypothesis asserts that P. Corneille and Molière could have collaborated, Molière providing drafts that P. Corneille would have later versified. If it were the case, we should observe a resemblance in lexicon in plays signed by Molière, but a dissemblance in rhymes' vocabulary, affixes, morphosyntax, and function words, that all should be closer to P. Corneille. Yet, nothing of the sort is observed for plays signed by Molière. In our final analysis, all the plays signed by Molière belong to the same cluster, very distinct from P. Corneille's plays, whichever the type of feature studied. In light of our results, we thus consider this first hypothesis disproved.

The second one states that Molière's plays would have entirely been written by one or more author(s) of his time, Molière only being the actor performing the lead role. If it were true, clusters mixing plays signed by P. Corneille (or another author) and Molière should appear. Yet again, our analyses show a clear-cut separation between all the plays by Molière and any other author studied, P. and T. Corneille included. This substantiates the claim that all of the plays signed by Molière have not been written by P. Corneille, nor one of the other authors studied here. The possibility remains that they are due to another author (or several authors with very similar styles) outside the scope of the major playwrights of the time, tested here, yet this hypothesis has never been put forward and seems implausible from a historical perspective.

Among the limitations of our study is the impossibility to include an analysis of plays in prose. P. Corneille did not write plays in prose, while Molière wrote many. For now, literature does not provide us with reliable tools to assess authorship in that case. This problem opens up venues for further investigation. It, however, does not undermine our confidence in our general result: It is quite unlikely that Molière would have written his own plays in verse but would have used a ghostwriter for his plays in prose. Advocates of the thesis according to which P. Corneille would be Molière's ghostwriter found that masterpieces in verse such as *Tartuffe*, *Le Misanthrope*, or *Amphitryon* were the plays that raised the most suspicion (*3*, *2*). Of course, this paper cannot be seen as a proof that another ghostwriter outside of this set could not have written all the plays under the name of Molière. Yet, it shows that these plays are very homogeneous in style and very likely written by a single individual, and that this person is not one of the authors whose plays have been analyzed here. In particular, it shows that, from any viewpoint adopted, it is very unlikely that P. Corneille or his brother Thomas would have been Molière's ghostwriters. As they were, after a century-old debate, the only option deemed plausible, these conclusions strongly substantiate the idea that Molière indeed wrote his own plays.

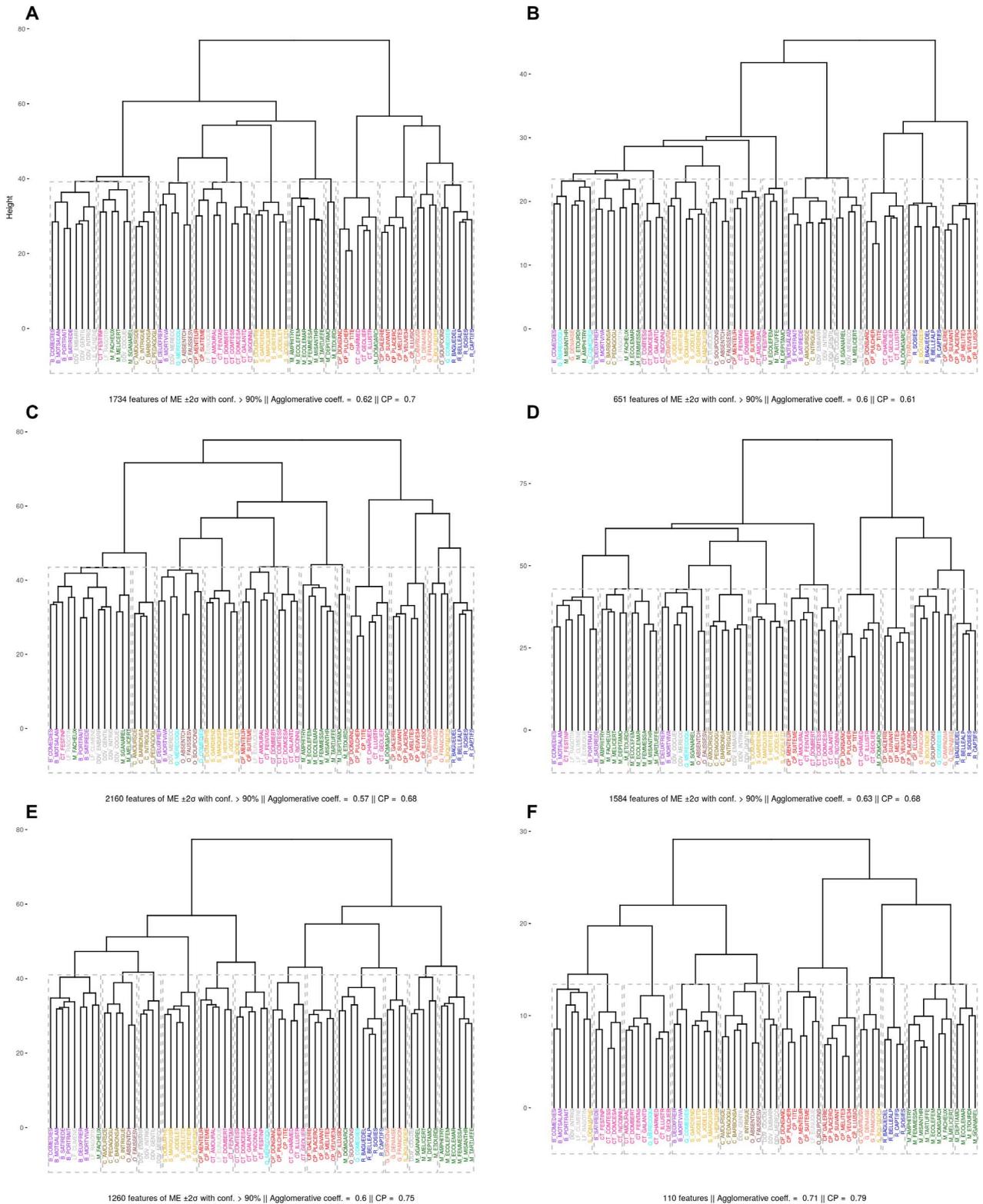

**Fig. 1. Dendrograms of agglomerative hierarchical clustering of the six feature sets, performed on the exploratory corpus (Ward's linkage criterion, Manhattan distance, z transformation, and vector length normalization), accompanied by the number of features selected, the agglomerative coefficient, and cluster purity for the exhibited clusters with respect to the alleged authors (see Materials and Methods).** Different sets of features are analyzed, from the most thematic to the most genre invariant: (**A**) lemma, (**B**) lemma in rhyme position, and (**C**) word forms, strongly related to the texts thematic contents; (**D**) affixes, (**E**) POS 3-grams, and (**F**) function words, which, in the current state of knowledge, are deemed to reflect most accurately the less conscious variations in individual style (B, Boursault; C, Chevalier; CP, Pierre Corneille; CT, Thomas Corneille; DDV, Donneau de Visé; G, Gillet de la Tessonerie; LF, La Fontaine; M, Molière; O, Ouville; Q, Quinault; R, Rotrou; S, Scarron).

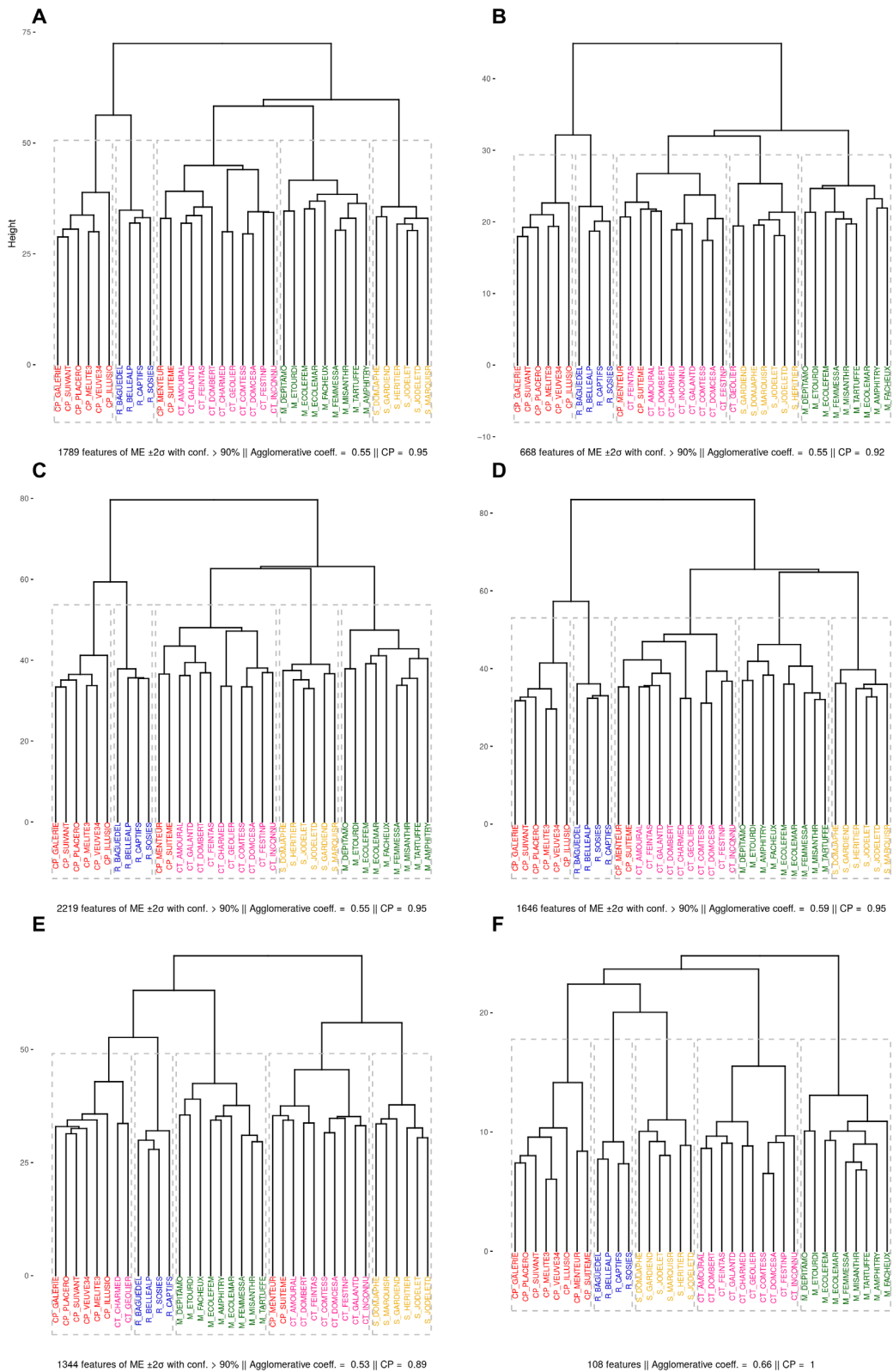

**Fig. 2. Dendrograms of agglomerative hierarchical clustering of the six feature sets** (Ward's linkage criterion, Manhattan distance, *z* transformation, and vector length normalization), accompanied by the number of features selected, the agglomerative coefficient, and cluster purity for the exhibited clusters with respect to the alleged authors (see Materials and Methods). Features analyzed: (**A**) lemma, (**B**) lemma in rhyme position, (**C**) word forms, (**D**) affixes, (**E**) POS 3-grams, and (**F**) function words. Despite variations in detail, each analysis shows clusters strongly or completely related to the putative authors (CP, Pierre Corneille; CT, Thomas Corneille; M, Molière; R, Rotrou; S, Scarron).

## MATERIALS AND METHODS
### Study design
Following Schöch (*13*), we used P. Fièvre's electronic corpus of Classical French Theatre (*30*). Genre was a strong potential bias: P. Corneille mostly signed tragedies, while Molière almost exclusively signed comedies. This would obviously be reflected on the lexical properties of the plays. What is more, Molière wrote many comedies in prose, but P. Corneille did not. This opposition between prose and verse could create artificial differences between the two sets of texts. We thus restricted our analysis to comedies in verse, the only genre where P. Corneille and Molière are supposed to have written a sufficient amount of plays.

Moreover, the stylometric analysis of 17th century theater poses specific difficulties. As previously stated, the relative weakness of authorial variation compared to the strength of intergeneric variation and the impact of generic constraints forces us to control and neutralize for other factors than authoriality.

For this reason, we restricted our studies to plays in verse, of the same genre, comedy. Moreover, comedy itself can be subdivided in subgenres, with noticeable thematic and stylistic difference. While shortest comedies can be marked by farcesque elements, the "grande comédie" borrows its structure and its meter to its nobler sister, tragedy. Most of the time, plays in one act are mostly light entertaining shows, while plays in five acts are more ambitious and serious. This will sediment at the turn of the century, giving a distinction between the *haut comique*, "devoted to an ideal world, knowing only virtue and error" (*32*), always written in alexandrines, and mostly in five acts, and less serious plays, often in less than three acts, and written in prose (*33*). Comparing plays of different lengths thus also means comparing plays belonging to different subgenres.

Variation in text length also has statistical implications, regarding both the minimum necessary size for reliable attribution and the effect of size variation in the analysis. If recent literature is obviously unanimous about the importance of the texts length (*34*) and the difficulty of working on short texts (*35*), length criteria deemed necessary to obtain reliable results in authorship attribution can vary, some authors seeming to achieve good results with texts under 1000 words (*36*, *37*), while recent systematic studies seem to advocate the study of more substantial texts (*38*).

To avoid this bias, we decided here to select only plays of at least 5000 words, which ensures a number of words strictly superior to any of the various criteria proposed. We also decided to exclude authors with less than three plays to avoid too wide differences in the sample sizes.

We used existing metadata from Fièvre's corpus (*30*) to select the comedies in verse, excluding all prose and mixed-form texts of appropriate length. An exception to this is Scarron's *Le Prince Corsaire*, *tragi-comédie*, which was (mistakenly) labeled as comédie in the metadata and which we excluded.

This selection process left us with 71 plays by 12 alleged authors: Edme Boursault (1638–1701), Chevalier (?–1674), P. Corneille (1606–1684), T. Corneille (1625–1709), Gilet de La Tessonerie (c. 1620–c. 1660), Jean de La Fontaine (1621–1695), Molière (1622–1673), Antoine Le Métel d'Ouville (1589–1655), Philippe Quinault (1635–1688), Jean de Rotrou (1609–1650), Paul Scarron (1610–1660), and Jean Donneau de Visé (1638–1710). The selected plays are displayed in the Supplementary Materials (tables S1 and S2).

Analyses performed on this corpus revealed the strength of subgeneric interactions, partly correlated to variations in length between the shortest plays, often in one act, and sometimes marked by farcesque elements, the "grandes comédies" describing the mores of the time and the heroic comedies, both mostly in five acts. Second, variation in text length also seems to have an impact on clustering results, even when length-based normalization is performed (such as the use of relative frequencies), because of nonlinearity in the relation between variable absolute frequency and text length (*39*, *40*). Third, as can be noticed from the exploratory results, variations in size in the authors' sample might also create some artefacts. Alleged authors for which a large number of plays are included seem to attract by chance some plays by underrepresented authors. Some isolates can be aggregated too early in the process, a random play in the corpus being by chance statistically close to its properties. This can artificially affect evaluation of the clustering such as cluster purity. When choosing a cut for the clustering adapted to the number of authors in the corpus, some hapaxes are not being clustered in a cluster of their own. Last, the presence of short plays has an effect on the reliability of some of the analyses. The number of features that can be analyzed with a given margin of error and confidence interval decreases with the size of documents. This was particularly true for the rhymes, because it considers only a small sample of the words of a given play.

In addition, the numerous cases of alleged plagiarism, doubtful attribution, possible collaborations, etc., interfere with the procedure and make the results less easily interpretable. They involve authors such as d'Ouville, Donneau de Visé, Quinault, or La Fontaine.

To increase the reliability of the results, we decided to build a subcorpus, eliminating both the shortest plays and the heroic comedies, to retain only plays mostly in five acts, from the subgroup of the major authors, P. and T. Corneille, Molière, Scarron, and Rotrou. It notably contains the grandes comédies written by these authors. This subcorpus is constituted of 37 plays (see the Supplementary Materials, table S1).

We excluded *Psyché*, a very rare case of declared collaborative authorship—this play being written by P. Corneille, Molière, and Quinault. Between the two available editions of P. Corneille's *La Veuve*, we chose the first edition of 1634 over the latter version of 1682 and we chose the edition of 1633 rather than 1682 of *Mélite*.

The length of the plays of this corpus is homogeneous, with only two outliers (Fig. 3A). The size of the samples by author is similar, yet still displays some differences, reflecting differences in their available production of comedies, with, on one hand, Corneille and Molière noticeably more prolific than Scarron and Rotrou. The length of the plays by author is again quite homogeneous, with the two outliers attributed to Molière (Fig. 3C).

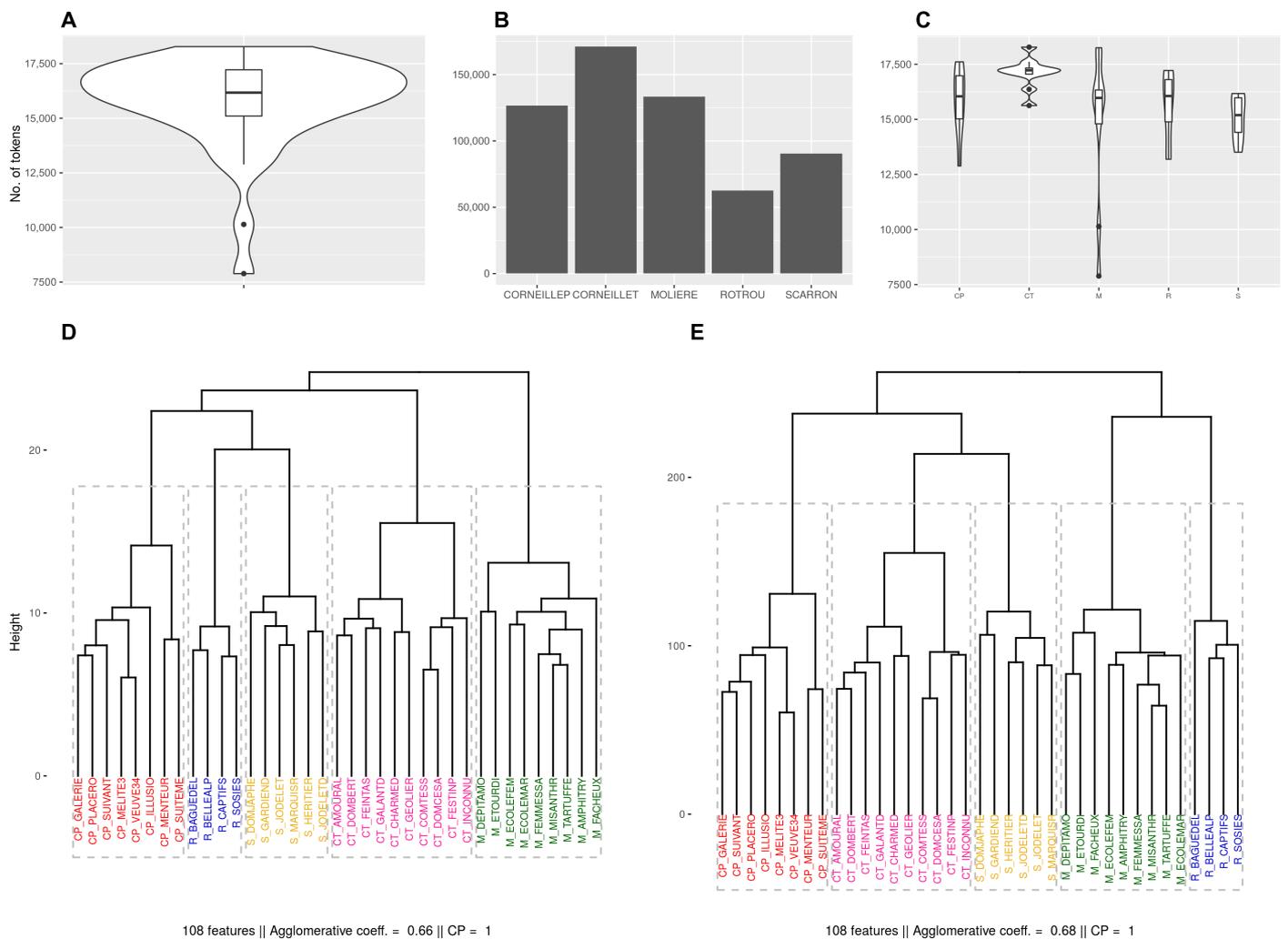

**Fig. 3. Distributions of the size in tokens of the texts and samples, and dendrograms of agglomerative hierarchical clustering on the function words (Ward's linkage criterion, Manhattan distance and MinMax metric, z transformation, accompanied by the number of features selected, the agglomerative coefficient, and cluster purity for the exhibited clusters with respect to the alleged authors (see Materials and Methods). (A)** The distribution of the length of texts in tokens for the final corpus for the initial corpus shows two outliers (too short texts) that were removed. **(B)** The size, per alleged author of the corpus, displays noticeable difference between authors due to differences in their production of comedies; in particular, the size of Rotrou and Scarron samples is relatively smaller. **(C)** In the corpus used for the final analyses, the chosen plays are relatively homogeneous in length (minimum, 7887; maximum, 18279) but still display some variation between authors. For cross-validation, we completed **(D)** the analysis on function words done with our main procedure, with **(E)** an analysis using the MinMax metric. The results have shown to be very similar, and the main clusters are identical in the set of their members.

The verses of the plays, with the exclusion of all surrounding materials, were extracted from Fièvre's digital edition (30). To be able to analyze the lemmata and POS 3-grams, we trained a neural lemmatizer (41) as well as a morphosyntactic tagger (42), specific to the French 17th century theater, until they achieved an accuracy of 98 and 97%, respectively. Technical details on text extraction and data preparation are given in the Supplementary Materials (section S1).

Six sets of features were selected for analysis: global lexicon, rhyme lexicon, word forms, affixes, morphosyntactic sequences, and the function words. In this corpus, the spelling was already normalized to the standard of Contemporary French, the most usual (yet questionable) practice in the edition of French 17th century texts. If this modernization is a loss of linguistic information, it was useful, in our case, to facilitate comparisons between authors, editions, or printers, without noise due to spelling. To eliminate other potential editorial or technical biases in the word counts, we further suppressed case distinctions and punctuation. For the analysis, we used a bag-of-word approach, common in stylometry (21).

For the global lexicon, the frequency of each lemma was computed, and all proper names were discarded. The same was done for the rhyme lexicon but restricted to the last lemma of each verse. The frequencies of word forms and affixes were also computed after the removal of proper names. Affixes were computed by extracting four types of 3-grams: "prefix," "suffix," "space prefix," and "space suffix" (26). For the word "_gloire_," this would yield the results "ˆglo," "ire$," "_gl," and "re_." 

For the morphosyntactic sequences, POS 3-grams (contiguous subsequences of length 3 of token POS tags) were extracted. The choice of 3-grams was consistent with the results of existing benchmarks (43, 44).

Last, the function words were examined. To eliminate the influence of the themes or characters evoked by the text, we worked on the 250 most frequent words, from which we removed all the remaining content-related words. We also removed personal pronouns and possessives, a deletion that has been shown to increase the accuracy of the results (45), because pronouns are suspected to be too heavily dependent on a text topic or genre (25). We kept verbal forms from *être* and *avoir*, which can be used as verbal auxiliaries in French. The set of all 103 retained function words is presented in the Supplementary Materials (table S4).

The selection of the most reliable and informative features for stylometric analysis is a question that has been the subject of many contributions, suggesting varying and sometimes contradictory cutoff levels for the features lists. To increase the reliability of the analyses, in a corpus with texts of varying length, we decided to select features based on the confidence level and margin of error that we could attain even for the smallest available sample in our corpus. The minimum sample size $n$ was calculated using the following formula (38)

$$n = \bar{p}(1 - \bar{p})\left(\frac{z}{e}\right)^2$$

where $\bar{p}$ is the feature mean probability in our corpus, used as an estimate of the population probability $\pi$, $z$ is the confidence level, and $e$ is the margin of error of the probability estimate. We set $z$ to obtain a confidence level above 90% and $e = 2\sigma$, where $\sigma$ is the feature standard deviation in the corpus.

For this equation to be valid, the features need to be normally distributed among samples. Because this is not always the case, we correct for normality by generating a mirror variable (38)

$$v\text{mirror}_{ji} = (\max_v + \min_v) - v_{ji}$$

where $v_j$ is the vector of the feature $j$, $\max_v$ and $\min_v$ are the maximum and minimum values in the feature vector $v_j$, and $v_{ji}$ is the probability estimate of the feature $j$ in a sample $i$. We then computed the arithmetic mean between this mirror value and the original value: This way, any overestimation by one value is compensated by an equivalent underestimation by its mirror value. We therefore obtained an unbiased estimate.

We retained only the features for which the minimum necessary sample size is superior to the shortest document. We cross-validated this procedure with a simple selection based on frequency rank, with different levels of cutoff.

## Statistical analysis

Many recent advances in authorship attribution have been made with supervised methods of attribution through machine learning. Jockers and Witten (46) have benchmarked several methods on the famous *Federalist Papers* corpus and have shown that regularized discriminant analysis and nearest shrunken centroids gave excellent results. The latter method has since then been used (47) or extended (48) in various papers.

In our case, however, supervised methods are not an option. Some theories to be tested imply that several alleged playwrights actually never wrote any plays. Defining a training corpus for those authors would thus be meaningless. This is why we chose to use an unsupervised method, regularly adopted in recent literature (35, 49, 50): hierarchical agglomerative clustering (for implementation, see section S2).

The notion of distance separating two textual entities is difficult to handle, as it is first and foremost a metaphor. A simple solution is to compute it with the classical Euclidean distance (5, 15). Yet, recent papers have advocated that the choice of the distance measure was a crucial problem in authorship attribution, which could markedly affect performance. Three benchmark studies, working on different sets of methods, have given some insights into which methods would be more suitable.

We thus used here a distance measure known for its efficiency for authorship attribution tasks (51, 52): Burrows' delta (53). A recent large-scale benchmark (21) notably showed that, combined with vector-length Euclidean normalization, this distance measure gave the best performance for French language (21), benchmarks using clustering methods such as hierarchical clustering with Ward's method. Burrows' delta computes the Manhattan distance between the $z$ scores of the coordinates

$$\Delta_{(AB)} = \sum_{i=1}^{n} \left| \frac{A_i - B_i}{\sigma_i} \right|$$

where $A_i$ and $B_i$ denote the frequency of a word $i$ in texts $A$ and $B$, respectively, and $\sigma_i$ is the variance of usage of word $i$. Following their experiments, we used with this method the relative word frequencies of the full texts (i.e., of varying lengths), without sampling.

Koppel and Winter (54) proposed to use a distance measure that was not studied by Evert *et al.* (21): the MinMax metric. Kestemont *et al.* (22) then showed in their benchmark that normalizing the term relative frequency by dividing it by its standard deviation (tfsd), similarly to Burrows' delta, outperformed the term-frequency/inverse-document-frequency normalization chosen by Koppel and Winter. According to them, this metric

$$\text{minmax}(\vec{A}, \vec{B}) = 1 - \left(\frac{\sum_{i=1}^{n} \min\left(tfsd(A_i), tfsd(B_i)\right)}{\sum_{i=1}^{n} \max\left(tfsd(A_i), tfsd(B_i)\right)}\right)$$

outperforms any other for authorship verification of text samples of the same size.

Without significant reasons to think that one or the other method would yield better results, we cross-validated the results shown above, using MinMax metric. Results obtained are very similar (Fig. 3, D and E).

We used Ward's aggregation method in both settings. This linkage criterion minimizes the total within-cluster variance (or maximizes the between-cluster variance, equivalently). Let $C_1$ and $C_2$ be two clusters, $G_1$ and $G_2$ their respective centroids, and $n_1$ and $n_2$ the number of individuals in the respective clusters. The distance $d$ between clusters, to be minimized, is defined by the equation

$$d^2(C_1, C_2) = \frac{n_1 \times n_2}{n_1 + n_2} d^2(G_1, G_2)$$

To evaluate the strength of the clustering structure, we computed its agglomerative coefficient. Let $d(i)$ be the dissimilarity between an observation $i$ and the first cluster it is merged with, and $n$ the number of observations. The agglomerative coefficient (AC) writes

$$AC = \frac{\sum_{i=1}^{n}(1 - d_i)}{n}$$

This coefficient grows mechanically with the number of observations, and thus cannot be used in our case to compare computations made for different corpora. Yet, it helps evaluating the strength of clusterings made for the same corpus but with different features.

### A look into individual features

To get a better grasp on our clusters' meaning, we can examine, for each analysis, which features are the most correlated with them.

To analyze the features most correlated to the clusters detected, we computed the correlation ratio $\eta^2$. For a quantitative variable $y$ and a qualitative variable with different levels $j$ and individuals $i$

$$\eta^2 = \frac{\sum_j \sum_i (y_{ij} - \bar{y}_{.j})^2}{\sum_j \sum_i (y_{ij} - \bar{y})^2}$$

the upper bar denoting arithmetic averaging.

This is particularly useful for the clusters resulting from the exploratory analysis, where generic interactions are suspected (Fig. 4 and Table 1).

The lemma *gloire* ("glory") is the most correlated feature with the clusters based on lemmas (Fig. 4A), the second and third most

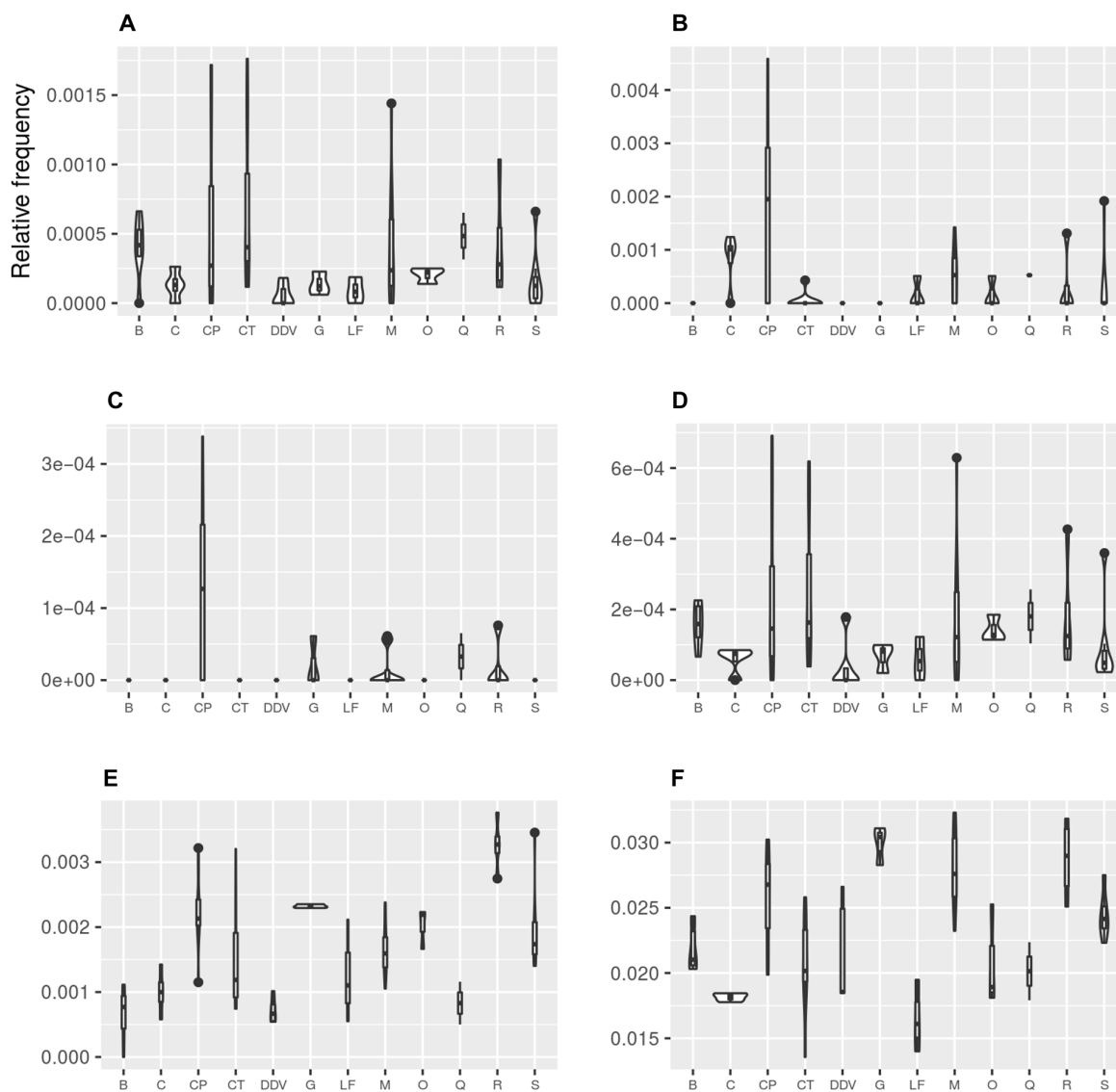

**Fig. 4. Distribution of the most discriminant features between clusters for each author.** (**A**) The lemma *gloire* ("glory"); (**B**) the lemma *contentement* ("contentment"), in rhyme position; (**C**) the form *contentements*; (**D**) the affix ˆ*glo*; (**E**) the POS sequence "DETdem ADJqua NOMcom" (demonstrative determiner, qualitative adjective, common noun); (**F**) the function word *et* ("and"). The feature with the strongest correlation, measured by $\eta^2$, with the five clusters of each analysis was selected (B, Boursault; C, Chevalier; CP, Pierre Corneille; CT, Thomas Corneille; DDV, Donneau de Visé; G, Gillet de la Tessonerie; LF, La Fontaine; M, Molière; O, Ouville; Q, Quinault; R, Rotrou; S, Scarron).

**Table 1. Features most correlated to the clusters detected, according to their correlation ratio $\eta^2$.**

| Feature | $\eta^2$ | $P$ | Feature | $\eta^2$ | $P$ | Feature | $\eta^2$ | $P$ |
|---|---|---|---|---|---|---|---|---|
| Lemma | | | Rhyme lemma | | | Forms | | |
| Gloire | 0.81 | $2.4 \times 10^{-17}$ | Contentement | 0.78 | $8.9 \times 10^{-16}$ | Contentements | 0.85 | $1.2 \times 10^{-20}$ |
| Négliger | 0.79 | $2.4 \times 10^{-16}$ | Affection | 0.78 | $3.0 \times 10^{-18}$ | Gloire | 0.82 | $3.4 \times 10^{-18}$ |
| Ha | 0.79 | $7.4 \times 10^{-16}$ | Gloire | 0.66 | $3.3 \times 10^{-10}$ | Ta | 0.80 | $1.4 \times 10^{-16}$ |
| Affection | 0.77 | $8.9 \times 10^{-15}$ | Courage | 0.65 | $6.7 \times 10^{-10}$ | Ha | 0.79 | $3.8 \times 10^{-16}$ |
| Contentement | 0.76 | $2.6 \times 10^{-14}$ | Sein | 0.59 | $4.2 \times 10^{-8}$ | Servage | 0.73 | $3.1 \times 10^{-13}$ |
| Bref | 0.75 | $8.8 \times 10^{-14}$ | Cavalier | 0.58 | $1.0 \times 10^{-7}$ | Bref | 0.73 | $4.5 \times 10^{-13}$ |
| Ton | 0.74 | $3.0 \times 10^{-13}$ | Maîtresse | 0.56 | $2.7 \times 10^{-7}$ | Affections | 0.72 | $1.1 \times 10^{-12}$ |
| Cela | 0.72 | $1.1 \times 10^{-12}$ | Négliger | 0.56 | $3.1 \times 10^{-7}$ | Tes | 0.71 | $5.0 \times 10^{-12}$ |
| Manière | 0.68 | $6.6 \times 10^{-11}$ | Éclat | 0.56 | $3.9 \times 10^{-7}$ | Pas | 0.70 | $1.3 \times 10^{-11}$ |
| De_le | 0.68 | $7.3 \times 10^{-11}$ | État | 0.55 | $4.7 \times 10^{-7}$ | Indigne | 0.70 | $1.4 \times 10^{-11}$ |
| Pas | 0.67 | $1.3 \times 10^{-10}$ | Envie | 0.55 | $5.5 \times 10^{-7}$ | Illustre | 0.69 | $2.8 \times 10^{-11}$ |
| Éclat | 0.66 | $2.6 \times 10^{-10}$ | Élément | 0.55 | $6.1 \times 10^{-7}$ | Dedans | 0.68 | $6.7 \times 10^{-11}$ |
| Dedans | 0.66 | $2.6 \times 10^{-10}$ | Souci | 0.53 | $2.1 \times 10^{-6}$ | Cela | 0.67 | $8.6 \times 10^{-11}$ |
| Illustre | 0.66 | $3.4 \times 10^{-10}$ | Enrager | 0.53 | $2.1 \times 10^{-6}$ | Éclat | 0.67 | $1.3 \times 10^{-10}$ |
| Bien | 0.66 | $3.8 \times 10^{-10}$ | Loi | 0.52 | $2.7 \times 10^{-6}$ | Manière | 0.66 | $2.3 \times 10^{-10}$ |
| Indigne | 0.65 | $5.4 \times 10^{-10}$ | Dieu | 0.52 | $3.3 \times 10^{-6}$ | Bien | 0.66 | $4.1 \times 10^{-10}$ |
| Être | 0.65 | $8.6 \times 10^{-10}$ | Cela | 0.52 | $3.5 \times 10^{-6}$ | Objets | 0.65 | $4.4 \times 10^{-10}$ |
| À | 0.65 | $9.1 \times 10^{-10}$ | Dire | 0.51 | $3.8 \times 10^{-6}$ | Affection | 0.65 | $7.5 \times 10^{-10}$ |
| Soleil | 0.64 | $2.0 \times 10^{-9}$ | Visage | 0.51 | $4.9 \times 10^{-6}$ | Ose | 0.65 | $8.5 \times 10^{-10}$ |
| Hé | 0.63 | $2.4 \times 10^{-9}$ | Dépit | 0.50 | $6.5 \times 10^{-6}$ | Transport | 0.63 | $2.2 \times 10^{-9}$ |
| Affixes | | | POS 3-gr | | | Function words | | |
| ^Glo | 0.77 | $5.3 \times 10^{-15}$ | DETdem.ADJqua.NOMcom | 0.67 | $9.0 \times 10^{-12}$ | Et | 0.71 | $2.9 \times 10^{-12}$ |
| Ha_ | 0.77 | $5.5 \times 10^{-15}$ | PROper.VERcjg.ADVgen | 0.66 | $1.8 \times 10^{-11}$ | Des | 0.70 | $9.4 \times 10^{-12}$ |
| Ta_ | 0.77 | $5.9 \times 10^{-15}$ | VERinf.PROper.VERcjg | 0.65 | $6.4 \times 10^{-11}$ | C' | 0.65 | $5.5 \times 10^{-10}$ |
| ^Dés | 0.74 | $2.8 \times 10^{-13}$ | PRE.DETpos.NOMcom | 0.64 | $8.7 \times 10^{-11}$ | Oui | 0.65 | $7.2 \times 10^{-10}$ |
| _Gl | 0.72 | $1.9 \times 10^{-12}$ | PRE.NOMcom.PROper | 0.64 | $1.0 \times 10^{-10}$ | De | 0.64 | $1.3 \times 10^{-9}$ |
| Et_ | 0.71 | $2.8 \times 10^{-12}$ | NOMcom.NOMpro.NOMpro | 0.64 | $1.4 \times 10^{-10}$ | Est | 0.64 | $1.7 \times 10^{-9}$ |
| _Dé | 0.71 | $3.3 \times 10^{-12}$ | NOMcom.PROrel.VERcjg | 0.63 | $2.7 \times 10^{-10}$ | Plus | 0.64 | $2.0 \times 10^{-9}$ |
| Ela$ | 0.71 | $4.0 \times 10^{-12}$ | NOMcom.CONsub.DETpos | 0.63 | $2.8 \times 10^{-10}$ | Voilà | 0.63 | $3.3 \times 10^{-9}$ |
| _L' | 0.70 | $9.7 \times 10^{-12}$ | DETpos.NOMcom.PRE | 0.63 | $3.0 \times 10^{-10}$ | En | 0.60 | $1.7 \times 10^{-8}$ |
| L'_ | 0.70 | $9.7 \times 10^{-12}$ | PROind.PRE.NOMcom | 0.62 | $4.2 \times 10^{-10}$ | S' | 0.60 | $2.2 \times 10^{-8}$ |
| _Bi | 0.69 | $1.6 \times 10^{-11}$ | CONsub.DETpos.NOMcom | 0.62 | $6.2 \times 10^{-10}$ | Ou | 0.60 | $2.5 \times 10^{-8}$ |
| ^Bie | 0.69 | $1.7 \times 10^{-11}$ | DETpos.NOMcom.ADVneg | 0.61 | $7.9 \times 10^{-10}$ | À | 0.59 | $5.8 \times 10^{-8}$ |
| _Ta | 0.69 | $2.6 \times 10^{-11}$ | PROper.VERinf.PROper | 0.60 | $1.9 \times 10^{-9}$ | Cela | 0.58 | $7.0 \times 10^{-8}$ |
| Ang$ | 0.69 | $2.8 \times 10^{-11}$ | PROper.PROadv.VERcjg | 0.60 | $1.9 \times 10^{-9}$ | Suis | 0.58 | $7.8 \times 10^{-8}$ |

*continued on next page*

| Feature | $\eta^2$ | P | Feature | $\eta^2$ | P | Feature | $\eta^2$ | P |
|---|---|---|---|---|---|---|---|---|
| Lemma | | | Rhyme lemma | | | | Forms | |
| Ng_ | 0.68 | $4.1 \times 10^{-11}$ | NOMcom.CONcoo.DETpos | 0.59 | $3.3 \times 10^{-9}$ | Encor | 0.58 | $8.1 \times 10^{-8}$ |
| Lat$ | 0.67 | $1.3 \times 10^{-10}$ | NOMcom.PRE.DETdef.NOMcom | 0.59 | $3.7 \times 10^{-9}$ | Ces | 0.56 | $2.2 \times 10^{-7}$ |
| _Et | 0.66 | $2.5 \times 10^{-10}$ | DETdem.NOMcom.ADJqua | 0.59 | $4.5 \times 10^{-9}$ | Encore | 0.56 | $2.5 \times 10^{-7}$ |
| _Je | 0.66 | $3.4 \times 10^{-10}$ | NOMcom.PRE.DETpos | 0.58 | $9.3 \times 10^{-9}$ | Non | 0.56 | $3.4 \times 10^{-7}$ |
| _À | 0.64 | $1.2 \times 10^{-9}$ | NOMcom.CONcoo.PRE | 0.58 | $1.1 \times 10^{-8}$ | Tous | 0.55 | $4.0 \times 10^{-7}$ |
| À_ | 0.64 | $1.2 \times 10^{-9}$ | PROadv.VERcjg.DETdef | 0.57 | $1.4 \times 10^{-8}$ | Enfin | 0.55 | $4.2 \times 10^{-7}$ |

correlated feature for word forms and rhymes (Table 1), in addition to the "affix" ʌ̂glo (Fig. 4D). The recurrence of this lemma in certain texts could well come from the topic of the plays and not from an author's style. Looking at the distribution of this feature in texts of the same alleged author (Fig. 4, A and D), we observe that, although the median value does not vary much between authors, the frequency varies strongly between texts, with several outliers, such as Molière's *Dom Garcie de Navarre*, P. Corneille's *Don Sanche d'Aragon*, and T. Corneille's *Illustres ennemis*. Observing this type of feature consolidates the hypothesis that the clusters are partially constituted by generic (e.g., heroic comedies or *comédies héroïques*) more than authorial criteria. Many lexicon-related features can be suspected of thematic or generic interference, being distinctive of a higher ("illustre," "indigne," and "éclat"; i.e., illustrious, unworthy, and luster) or plainer language (interjections, such as "ha"). Some other terms are related to the love intrigues prominent in many plays ["affection," "maîtresse," and "transport"; i.e., contentment, affection, mistress, and (amorous) transports]. Terms like "contentement(s)" in rhyme position or as a word form (Fig. 4, B and C) can be both thematic and authorial: The use of this feature seems characteristic of P. Corneille, yet with very important variations between his plays.

In contrast, looking at the distribution of more authorial features, such as the POS sequence "demonstrative determiner, qualificative adjective, and common noun" or the function word *et* (Fig. 4, E and F), reveals distributions that are closer to normality for each author but with stronger differences of median between them. The less thematic nature of these features makes them harder to perceive, to identify intuitively, or to interpret in other than authorial terms.

### Control corpus
Because of the doubts about authorship at Molière's time, we cannot use a sample of our corpus of interest to test our method's accuracy. There would be no "ground truth" everyone would agree on. To allow comparison and to assess our method's performance, we thus build a control corpus consisting of comedies in verse written right after P. Corneille's (and Molière's) death. We select them using exactly the same criteria as our final corpus and apply the same methods to them (Fig. 5).

Clustering results obtained on word forms, affixes, and morphosyntactic sequences (Fig. 5, C to E) attribute with a 100% success rate the plays to their author. Lemmas and function words analysis (Fig. 5, A and F) correctly attributes 94% of the plays. The only error comes from two plays by Dancourt located in the cluster of Regnard's plays (*La Métempsycose des amours* and *Sancho Pança, gouverneur*). This confusion could very well be an artefact but could also have deeper grounds. Dancourt's wife, Marie-Thérèse Le Noir de la Thorillière, was a famous actress, who created leading roles of many plays by Regnard present in our corpus. In particular, his wife and his daughter were part of the initial cast of Regnard's *Démocrite*, the play closer to the two dislocated plays of Dancourt. But there is more: Dancourt himself, being also an actor, took on the leading role of this play, after the initial actor (M. Poisson) failed to meet the expectations of the audience (30). Moreover, the second closest Regnard's play, the *Folies amoureuses*, starts by a prologue in which actors playing themselves discuss the merits of the play, including "Monsieur Dancour." These clues are deserving further investigation but may point to a form of collaborative authorship.

Rhymes' analysis (Fig. 5B) exhibits clusters that, while still being mainly authorial (cluster purity at 87%), offer a somewhat less clear situation that can perhaps be partly explained by statistical reasons. The sample of rhymes is significantly smaller than the sample of all the words in a text, limiting the number of features that can be used with sufficient reliability.

### Robustness checks
To monitor our clustering's performance and evaluate its robustness to variation in the selection of features, we repeat the previous analyses for each feature set, with a selection based only on total frequency. We perform our clustering with different levels of selection, ranging from the 1% most frequent features to all features. Each time, we compute two indexes: cluster purity with respect to a clustering by alleged author and cluster purity in comparison to the clusters shown in Figs. 2 and 5.

Cluster purity (CP) gives the percent of the total number of plays that were classified "correctly." Let $N$ be the number of individuals (plays in our case), $k$ the number of clusters, $c_i$ an observed cluster, and $t_j$ a cluster of the "ground truth" set of classes. Cluster purity then writes

$$CP = \frac{1}{N} \sum_{i=1}^{k} \max_j |c_i \cap t_j|$$

Results for both the control corpus and the final corpus (Table 2) are robust to significant variations in the selection of features. In particular, affixes prove to be the most robust feature set. These results also demonstrate the capacity of the selection procedure we used (retaining features of maximum margin of error ±2σ with confidence >90%) to attain the best performance level for a given feature set, with only one exception.

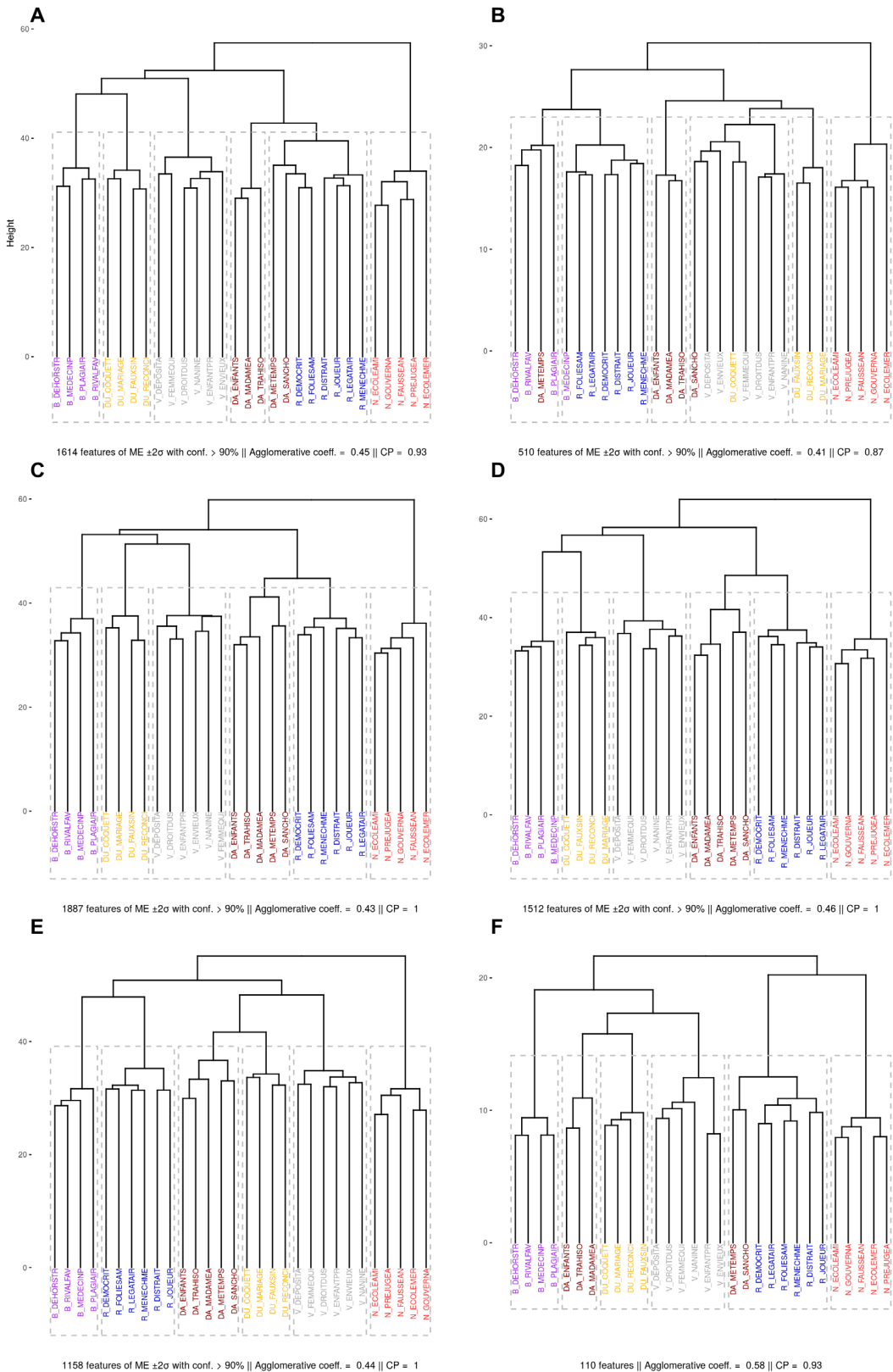

**Fig. 5. Dendrograms of agglomerative hierarchical clustering of the six feature sets of the control corpus (Ward's linkage criterion, Manhattan distance, *z* transformation, and vector length normalization), accompanied by the number of features selected, the agglomerative coefficient, and cluster purity for the exhibited clusters with respect to the alleged authors (see Materials and Methods).** Six sets of features are analyzed: (**A**) lemma and (**B**) lemma in rhyme position, (**C**) word forms and (**D**) affixes, (**E**) POS 3-grams, and (**F**) function words. B, Boissy; DA, Dancourt; DU, Dufresny; N, Nivelle; R, Regnard; and V, Voltaire.

**Table 2. Evaluation of the robustness of the clustering results.** For each feature set, the clustering is performed with level of selection from the 1% most frequent to all (MF); the number of features is given (*N*), as well as the cluster purity with respect to alleged authors (P-A) and with respect to the analysis obtained through our reference selection procedure (P-R). The last line, in italics, shows the results obtained with the reference selection procedure (RS), as shown in the dendrograms of Figs. 1, 2, and 5. Results with cluster purity above 0.9 are shown in bold. When P-A = P-R for all frequency cutoff thresholds, the reference selection procedure can be considered as optimal.

| | | | | **Control corpus** | | | | | | | |
|---|---|---|---|---|---|---|---|---|---|---|---|
| MF | *N* | P-A | P-R | MF | *N* | P-A | P-R | MF | *N* | P-A | P-R |
| **Lemmas** | | | | **Word forms** | | | | **POS 3-gr.** | | | |
| 1% | 80 | **0.90** | **0.97** | 1% | 155 | **0.90** | **0.90** | 1% | 95 | **0.90** | **0.90** |
| 10% | 795 | **1** | **0.93** | 10% | 1546 | **1** | **1** | 10% | 948 | **0.97** | **0.97** |
| 25% | 1986 | **1** | **0.93** | 25% | 3865 | **1** | **1** | 25% | 2369 | **0.97** | **0.97** |
| 50% | 3971 | **0.90** | **0.97** | 50% | 7730 | 0.83 | 0.83 | 50% | 4738 | 0.80 | 0.80 |
| 75% | 5998 | **0.90** | **0.97** | 75% | 11704 | 0.57 | 0.57 | 75% | 7117 | 0.73 | 0.73 |
| 100% | 7941 | **0.97** | **0.97** | 100% | 15457 | 0.80 | 0.80 | 100% | 9476 | 0.87 | 0.87 |
| *RS* | *1614* | *0.93* | – | *RS* | *1887* | *1* | – | *RS* | *1158* | *1* | – |
| **Lemmas in rhyme position** | | | | **Affixes** | | | | **Function words** | | | |
| 1% | 55 | 0.83 | 0.87 | 1% | 32 | 0.77 | 0.77 | 1% | 2 | 0.50 | 0.57 |
| 10% | 543 | **0.90** | **0.93** | 10% | 317 | **0.93** | **0.93** | 10% | 11 | **0.90** | **0.90** |
| 25% | 1358 | **0.90** | 0.77 | 25% | 792 | **1** | **1** | 25% | 28 | 0.87 | **0.93** |
| 50% | 2715 | 0.77 | 0.80 | 50% | 1583 | **1** | **1** | 50% | 55 | **0.90** | **0.97** |
| 75% | 4114 | 0.70 | 0.70 | 75% | 2374 | **1** | **1** | 75% | 82 | **0.93** | **1** |
| 100% | 5429 | 0.73 | 0.73 | 100% | 3165 | **0.97** | **0.97** | 100% | 110 | **0.93** | **1** |
| *RS* | *510* | *0.87* | – | *RS* | *1512* | *1* | – | *RS* | *110* | *0.93* | – |
| | | | | **Main corpus (subgroup)** | | | | | | | |
| MF | *N* | P-A | P-R | MF | *N* | P-A | P-R | MF | *N* | P-A | P-R |
| **Lemmas** | | | | **Word forms** | | | | **POS 3-gr.** | | | |
| 1% | 88 | **0.92** | **0.97** | 1% | 176 | **0.95** | **1** | 1% | 103 | 0.89 | **0.95** |
| 10% | 879 | **0.95** | **1** | 10% | 1754 | **0.95** | **1** | 10% | 1025 | 0.86 | **0.97** |
| 25% | 2196 | **0.92** | **0.97** | 25% | 4385 | **0.95** | **1** | 25% | 2561 | 0.89 | **1** |
| 50% | 4391 | **0.92** | **0.97** | 50% | 8770 | **0.95** | **1** | 50% | 5122 | 0.81 | 0.81 |
| 75% | 6635 | 0.81 | 0.86 | 75% | 13228 | 0.70 | 0.76 | 75% | 7710 | 0.81 | 0.81 |
| 100% | 8781 | 0.73 | 0.78 | 100% | 17540 | 0.89 | **0.95** | 100% | 10243 | 0.86 | 0.81 |
| *RS* | *1789* | *0.95* | – | *RS* | *2219* | *0.95* | – | *RS* | *1344* | *0.89* | – |
| **Lemmas in rhyme position** | | | | **Affixes** | | | | **Function words** | | | |
| 1% | 58 | 0.81 | 0.86 | 1% | 33 | 0.89 | **0.95** | 1% | 2 | 0.59 | 0.59 |
| 10% | 573 | **0.92** | **0.95** | 10% | 326 | **0.95** | **1** | 10% | 11 | 0.62 | 0.62 |
| 25% | 1432 | **0.95** | **0.97** | 25% | 814 | **0.95** | **1** | 25% | 27 | 0.86 | 0.86 |
| 50% | 2864 | **0.95** | **0.95** | 50% | 1627 | **0.95** | **1** | 50% | 54 | **1** | **1** |
| 75% | 4298 | 0.76 | 0.76 | 75% | 2440 | **0.92** | **0.97** | 75% | 81 | **1** | **1** |
| 100% | 5728 | 0.68 | 0.73 | 100% | 3253 | **0.92** | **0.97** | 100% | 108 | **1** | **1** |
| *RS* | *668* | *0.92* | – | *RS* | *1646* | *0.95* | – | *RS* | *108* | *1* | – |

## SUPPLEMENTARY MATERIALS

Supplementary material for this article is available at http://advances.sciencemag.org/cgi/content/full/5/11/eaax5489/DC1

Section S1. Data preparation
Section S2. Data analysis implementation
Table S1. Plays used in the final (subgroup) analysis.
Table S2. Plays used only in the exploratory study.
Table S3. Plays of the control corpus.
Table S4. Function words used for the analyses.
Table S5. List of cluster members for each dendrogram shown in the main text.
Data file S1. Training corpus for the lemmatizer and POS tagger, in tsv format, with the trained models.
Data file S2. Automatically labeled corpora in xml format, with import scripts.
Data file S3. Feature datasets and analysis scripts in csv, R, and RMarkdown formats.
References (55–58)

**Acknowledgments:** We wish to thank S. Gabay (University of Neuchâtel) for his contribution in the production of the labeled corpus used to train the lemmatizer and POS tagger. We are also grateful to J. K. Ward (Sorbonne University) and J. Randon-Furling (Panthéon-Sorbonne University) for their careful reading of the manuscript. Errors remain our own. **Funding:** This research was partially funded through the Université PSL E-Philologie grant (2014-2017) and benefited from the developments of the LAKME project (Linguistically Annotated Corpora Using Machine Learning Techniques, 2015-2018; Université PSL). **Authors contributions:** Both authors contributed equally to all aspects of this paper. **Competing interests:** The authors declare that they have no competing interests. **Data and materials availability:** All data needed to evaluate the conclusions in the paper are present in the paper and/or the Supplementary Materials. Additional data related to this paper may be requested from the authors.

Submitted 2 April 2019
Accepted 4 October 2019
Published 27 November 2019
10.1126/sciadv.aax5489

**Citation:** F. Cafiero, J.-B. Camps, Why Molière most likely did write his plays. *Sci. Adv.* 5, eaax5489 (2019).

# Science Advances

## Why Molière most likely did write his plays

Florian Cafiero and Jean-Baptiste Camps

| | |
|---|---|
| ARTICLE TOOLS | http://advances.sciencemag.org/content/5/11/eaax5489 |
| SUPPLEMENTARY MATERIALS | http://advances.sciencemag.org/content/suppl/2019/11/21/5.11.eaax5489.DC1 |
| REFERENCES | This article cites 21 articles, 0 of which you can access for free<br>http://advances.sciencemag.org/content/5/11/eaax5489#BIBL |
| PERMISSIONS | http://www.sciencemag.org/help/reprints-and-permissions |

Use of this article is subject to the Terms of Service